\documentclass{article}
\usepackage{amsmath,amssymb,amsfonts}  
\usepackage{graphicx}                   
\usepackage{hyperref}                   
\usepackage{cite}                       

\usepackage{algorithm}
\usepackage{algorithmic}

\title{Differential Informed Auto-Encoder}
\author{Zhang Jinrui\thanks{alternative email:zhangjr1022@mails.jlu.edu.cn} \\ \texttt{jerryzhang40@gmail.com}}

\date{20241021}  

\begin{document}

\maketitle

\begin{abstract}
    In this article, an encoder was trained to obtain the inner
    structure of the original data by obtain a differential equations.
    A decoder was trained to resample the original data domain, to generate new data that obey the differential structure of the original data using the physics-informed neural network\cite[PINN]{raissi2017physics}.
\end{abstract}

\section{Introduction}
If the physics formula was obtained in the form of differential equations,
a physics-informed neural network can be built to solve it numerically
on a global scale\cite[PINN]{raissi2017physics}.This process could be seen
as a decoder in a way that takes a sample point in the domain of
the partial differential equations, and solve it to get the corresponding
output of each input point. If only a small and random amount of
training data was obtained, to re-sample from the domain, we need
to obtain the differential relationship of the data. This process
could be viewed as an encoder that encodes the inner structure of
the original data. And the decoder decode it by solving the
differential equations.

\section{Methodology}
\subsection{first approach}
The first idea is simple. For a one-variable function $u(t)$,
define a second-order differential equation
in its general form $(\forall t)(F(\frac{d^2u}{{du}^2},\frac{du}{dt},u)=0)$.

The data of the function $u(t)$ are given in tuples denote as
$(T,U)_i\equiv(T_i,U_i)$. And it is natural to denote the differentials
by $U^{t}_{i}$ and $U^{tt}_{i}$. There are several methods to compute
these two differentials, including just using the definition of
the derivative. In this article, local PCA are compute to obtain
these differentials. Local PCA means finding the nearest K neighbors
of a given point, which K is a hyper parameter, and performing PCA
on these points close to each other to get the principal direction.
The slope of this direction is the derivative $U^{t}$ in general.
Repeat this process on $(T,U^{t})$ to obtain $U^{tt}$

Create a FCN denote as $f$ to represent
$F(\frac{d^2u}{{du}^2},\frac{du}{dt},u)$
$F$ to be $0$ at every data points and to be $1$ all elsewhere is wanted.

To achieve these requirements, we evaluate $f$ at all the data points,
and train the network to evaluate these points to $0$. Then randomly sample
the points of $\mathbb{R}^{3}$ and train these points to be $1$ Algorithm\ref{algor:1}.

\begin{algorithm}
    \caption{$f$ trainer}\label{algor:1}
    \begin{algorithmic}[1]
        \REQUIRE Input parameters $f,T_i, U_i, U^{t}_{i}, U^{tt}_{i}$
        \STATE Initialize $f$ randomly
        \REPEAT
        \STATE $F_i \leftarrow f(U_i,U^{t}_{i},U^{tt}_{i})$
        \STATE $RAND_i \leftarrow$ randomly sample in $\mathbb{R}^{3}$
        \STATE $R_i \leftarrow f(RAND_i)$
        \STATE $L \leftarrow meanSquareError(F_i,0)+0.1*meanSquareError(R_i,1)$
        \STATE back Propagation against $L$ to optimize $f$
        \UNTIL {$L$ meets requirement}
        \RETURN $f$
    \end{algorithmic}
\end{algorithm}

Once The $f$ was obtained, we can perform PINN as a decoder
to generate new data.

The experiment code for the pictures in Results can
be run by a Python program in Github\cite[deSineTasks]{firstApproachGithubProject}
The requirement environment may be installed using\cite[reqs]{Envreqs}

\subsection{approach with linear assumption}
\begin{figure}[ht!]
    \centering
    \includegraphics[width=1.0\textwidth]{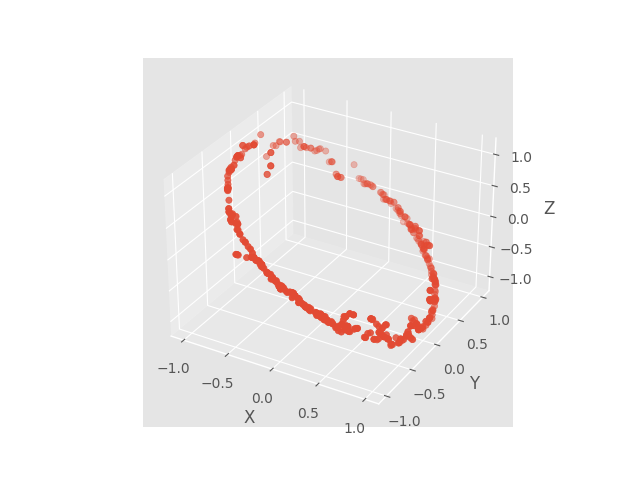}
    \caption{vectors $\left[\begin{matrix}{U}&{U^{t}}&{U^{tt}}\end{matrix}\right]$ sit on 1-dimensional manifold}
    \label{one:dimensionalmani}
\end{figure}

For a small randomly sampled data set, the data points
in the differential vector space
$\left[\begin{matrix}{U}&{U^{t}}&{U^{tt}}\end{matrix}\right]$
are a low-dimensional manifold. With only one equation to satisfy, the dimension of the manifold in the latent space would be exactly one less than the whole space dimension. In the second-order
differential equation case, the manifold has to be a two-dimensional
manifold. As the pictures show in Figure\ref{one:dimensionalmani},
if we only have some of the experimental data input, with
the algorithm in the first approach, we would set all the
values in the one-dimensional manifold to $0$ but all elsewhere
to $0$ which is only one specify solution with the same
initial condition as the input data.

To have more generalization ability, sacrifice
is necessarily taken with some of the flexibility
to be able to learn all kind of weird data structures,
but assume that we are learning a linear equation at the first place.
In this specific case, the ring shape data points in Figure\ref{one:dimensionalmani}
need to be treated as a plane crossing through the ring span.

As long as the data are on the spanned plane. we will have
the same differential structure as the data set.
With this assumption, basically only a specific clean data
set are required for only one initial condition.

Obtaining this result is also very straightforward.
Perform all the methods that have been used to compute the differential vector
$\left[\begin{matrix}{U}&{U^{t}}&{U^{tt}}\end{matrix}\right]$
in the first approach. Either directly get the derivatives or
perform other methods.

After $U_i, U^{t}_{i}, U^{tt}_{i}$ are obtained, the
latent space of the differential relationships is obtained.
With the linear assumption, the equation is linear so all the
valid points sit on one same plane. Perform PCA directly on
these data points, then take the eigen vector $v$ of the
smallest singular value, the direction that needs to be reduced
has been found. Simply define the encoder function as
$f(x)=v \cdot x$ Algorithm\ref{algor:2}.

Then minimizing $f$ will reduce one dimension linearly
from the latent differential space. The function $f$ has
done the job originally constrained by the differential equation.

This method can find all the linear differential relationships, i.e. the linear differential equations,
from a single function from the entire solution function family.
\begin{algorithm}
    \caption{normal vector $v$ calculator}\label{algor:2}
    \begin{algorithmic}[1]
        \REQUIRE Input parameters $U_i, U^{t}_{i}, U^{tt}_{i}$
        \STATE stack $(U_i,U^{t}_{i},U^{tt}_{i})$ to latent vectors $LAT_{ij}$
        \STATE perform $PCA(LAT_{ij})$
        \STATE $v \leftarrow$ the eigen vector of the smallest singular value
        \RETURN $v$
    \end{algorithmic}
\end{algorithm}

The experiment result can be obtained by Github
code \cite[deLinearTasksSine]{deLinearTasksSine}

\subsection{approach augmentation}
This method is a AutoEncoder we can also check the difference
betwen the input data and the data generate by the PINN in the
differential relationship constrain. This can view as a parameterize method.

If the Manifold Hypothesis hold, a high dimensional
dataset $x_\delta$ is a lower-dimensional manifold $\rho_d$ embedding in
the high-dimension space. Denote the dimension of the
original data $\Delta$, and denote the dimension of the
latent variable $D$. To parameterize the data manifold,
we need a decoder $\mathbb{R}^{D} \xrightarrow{f^{-1}} \mathbb{R}^{\Delta}$
, ${\rho}_d \mapsto y_\delta$. The jacobian of ${f^{-1}}$
denoted as $J({f^{-1}})=J^\delta_d=\frac{\partial y_\delta}{\partial \rho_d}$.
Denote the second-order jacobian as
$J^2({f^{-1}})=J^\delta_{d_1 d_2}=\frac{\partial^2 y_\delta}{\partial \rho_{d_1}\partial \rho_{d_2}}$.
For the sake of symbolic coherence, denote
$J^0({f^{-1}})=J^\delta$
To find a linear differential equation means to find
equation in the form of
$A_{d_1 d_2}J^\delta_{d_1 d_2}+A_{d}J^\delta_d+AJ^\delta=0$. The generate
equation of order $N$ could be written as
$\sum_{j=0}^{N} A_{d_1 d_2 ... d_j}J^\delta_{d_1 d_2 ... d_j}=0$.
As we only want to find the direction of the hyper plane,
the normal vector concatenate $(\forall j\leq N)A_{d_1 d_2 ... d_j}$
denoted as $V$, need to be normalized. To check this idea correctly, the $V$ is a
vector of dimension $\sum_{j=0}^{N} D^{j}$, so this algorithm
may be computaional heavy to some higher-order differential
relationship encode, or to some high latent dimension encode task.

For the other side of the AutoEncoder, we denote
the encoder as $\mathbb{R}^{\Delta} \xrightarrow{f} \mathbb{R}^{D}$,
$x_\delta \mapsto {\rho}_d$. The normal AutoEncoder
requires minimizing the mean square error between
$x_\delta, y_\delta$. In the differential informed
method, the
term $A_{d_1 d_2}J^\delta_{d_1 d_2}+A_{d}J^\delta_d+AJ^\delta$ also needs to be optimized to zero Algorithm\ref{algor:3}.
\begin{algorithm}
    \caption{normal vector $v$ calculator}\label{algor:3}
    \begin{algorithmic}[1]
        \REQUIRE Input parameters $f, f^{-1}, x_\delta ,(\forall j\leq N)A_{d_1 d_2 ... d_j}$
        \STATE Initialize $f, f^{-1}$ randomly
        \REPEAT
        \STATE $\rho_d \leftarrow f(x_\delta)$
        \STATE $y_\delta \leftarrow f^{-1}(\rho_d)$
        \STATE $L \leftarrow meanSquareError(x_\delta,y_\delta)$
        \STATE back Propagation against $L$ to optimize $f, f^{-1}$
        \UNTIL {$L$ meets requirement}
        \RETURN $f$
        \REPEAT
        \STATE $\rho_d \leftarrow f(x_\delta)$
        \STATE $y_\delta \leftarrow f^{-1}(\rho_d)$
        \STATE $(\forall j\leq N)A_{d_1 d_2 ... d_j} \leftarrow$ calculate the Jacobian of each order
        \STATE $L \leftarrow mse(x_\delta,y_\delta)+mse(\sum_{j=0}^{N} A_{d_1 d_2 ... d_j}J^\delta_{d_1 d_2 ... d_j},0)$
        \STATE back Propagation against $L$ to optimize $f, f^{-1},(\forall j\leq N)A_{d_1 d_2 ... d_j}$
        \UNTIL {$L$ meets requirement}
        \RETURN $f$
    \end{algorithmic}
\end{algorithm}

The result of the experiment can be obtained using the Github
code \cite[deLinearAugSine]{deLinearAugSine}

\section{Results}
\subsection{first approach}
\begin{figure}[ht!]
    \centering
    \begin{minipage}{0.45\textwidth}
        \centering
        \includegraphics[width=0.9\textwidth]{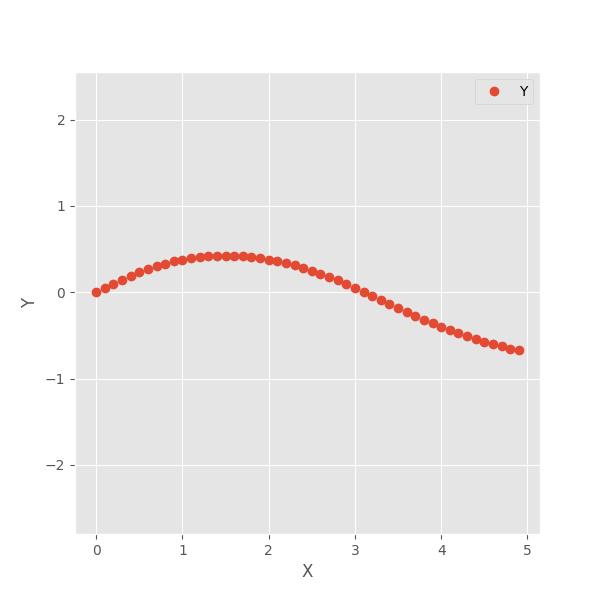} 
        \caption{$0.5*sin(x)$}
        \label{fig:fig1}
    \end{minipage}\hfill
    \begin{minipage}{0.45\textwidth}
        \centering
        \includegraphics[width=0.9\textwidth]{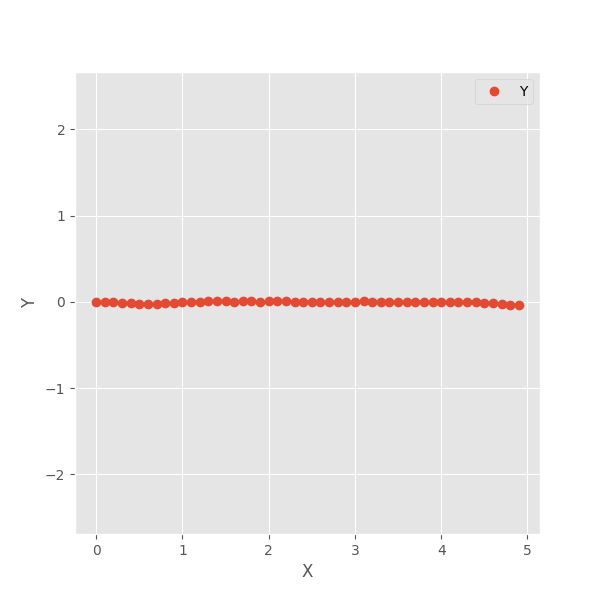} 
        \caption{$f$ errors}
    \end{minipage}
\end{figure}

Train the model on a pure $sin(x)$ and try to
get a result that satisfies the initial condition with
$U^{t}_0=0.5$ and $U_0=0.0$ to which the exact solution is $0.5*sin(x)$ would. The result is shown in Figure\ref{fig:fig1}.
\subsection{approach with linear assumption}
\begin{figure}[ht!]
    \centering
    \begin{minipage}{0.45\textwidth}
        \centering
        \includegraphics[width=0.9\textwidth]{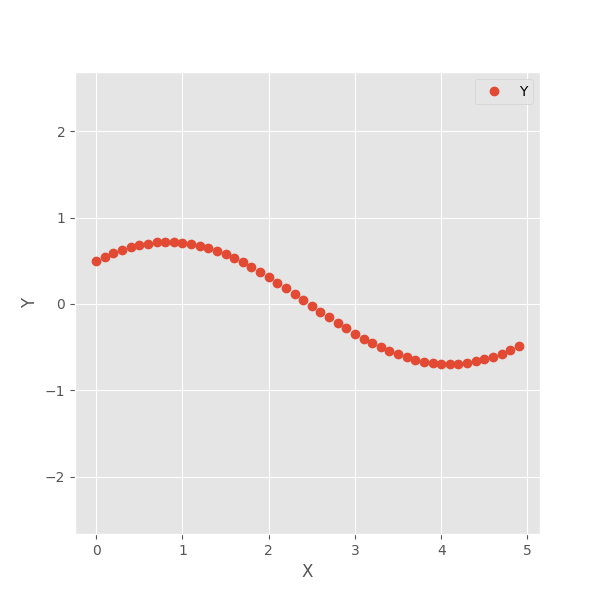} 
        \caption{$\frac{\sqrt{2}}{2}*sin(x+\frac{\pi}{4})$}
        \label{fig:fig2}
    \end{minipage}\hfill
    \begin{minipage}{0.45\textwidth}
        \centering
        \includegraphics[width=0.9\textwidth]{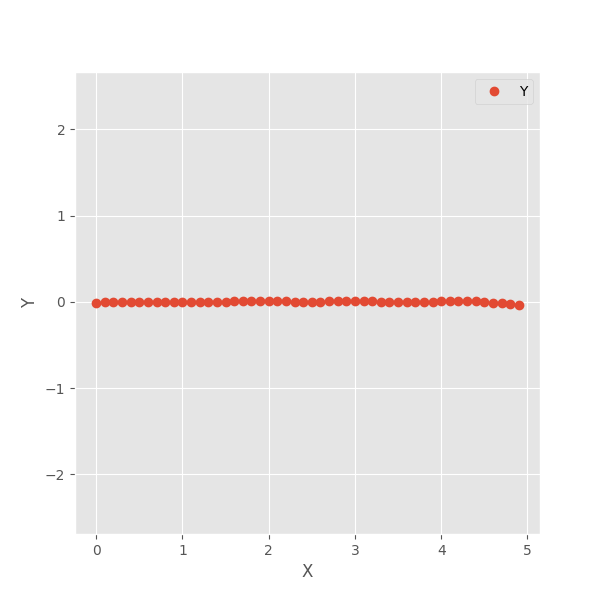} 
        \caption{$f$ errors}
    \end{minipage}
\end{figure}

Train the model on a pure $sin(x)$ and try to
get a result that satisfies the initial condition with
$U^{t}_0=0.5$ and $U_0=0.5$ in which the exact solution is $\frac{\sqrt{2}}{2}*sin(x+\frac{\pi}{4})$ would be
required output. The result is shown in Figure\ref{fig:fig2}.
\subsection{approach augmentation}
\begin{figure}[ht!]
    \centering
    \begin{minipage}{0.45\textwidth}
        \centering
        \includegraphics[width=0.9\textwidth]{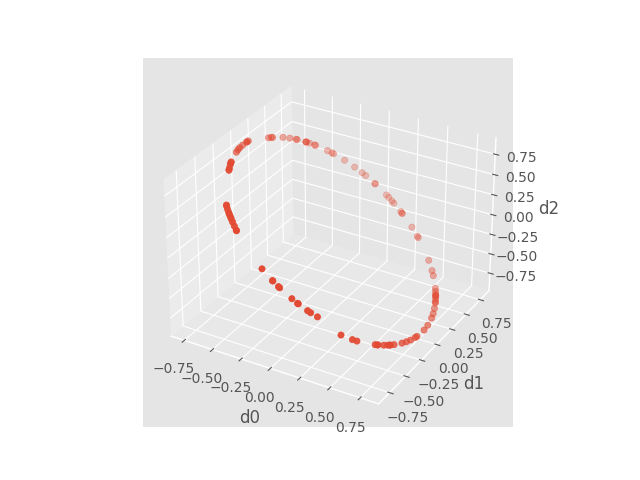} 
        \caption{first component latent space of $y^\delta(\rho)$}
        \label{fig:fig3}
    \end{minipage}\hfill
    \begin{minipage}{0.45\textwidth}
        \centering
        \includegraphics[width=0.9\textwidth]{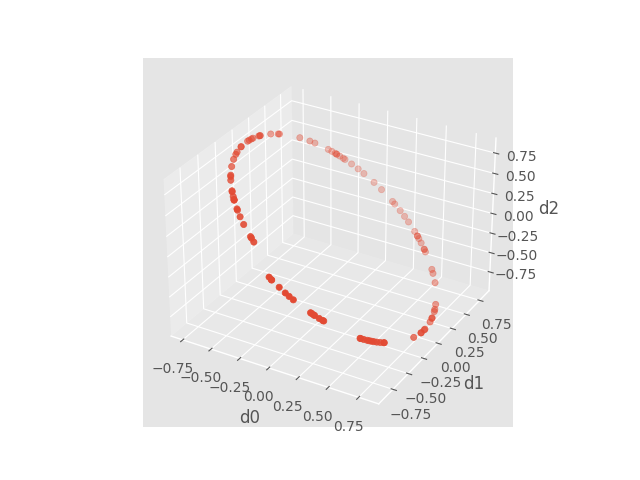} 
        \caption{second component latent space of $y^\delta(\rho)$}
    \end{minipage}
\end{figure}

\begin{figure}[ht!]
    \centering
    \begin{minipage}{0.45\textwidth}
        \centering
        \includegraphics[width=0.9\textwidth]{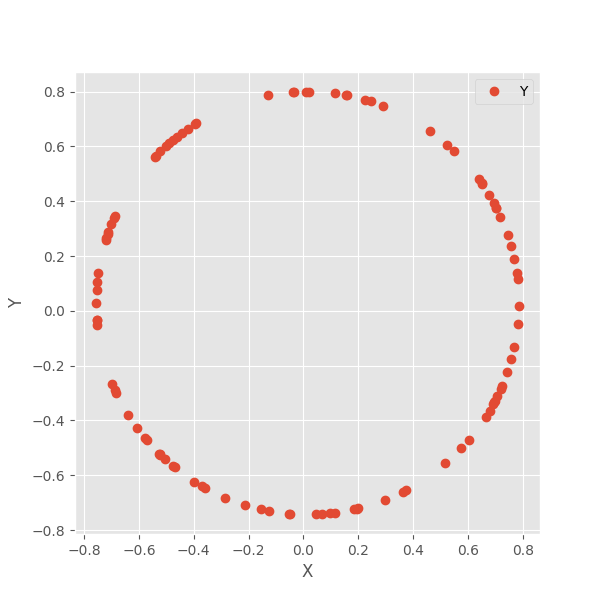} 
        \caption{original data $x_\delta$}
        \label{fig:fig4}
    \end{minipage}\hfill
    \begin{minipage}{0.45\textwidth}
        \centering
        \includegraphics[width=0.9\textwidth]{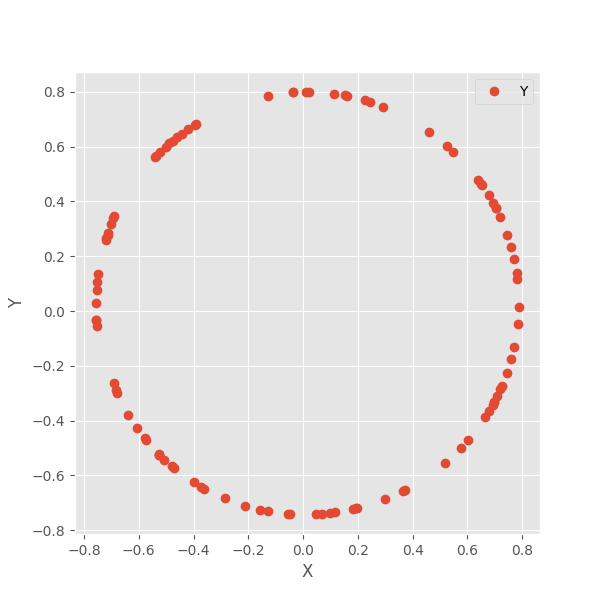} 
        \caption{regenerate data $y_\delta$}
    \end{minipage}
\end{figure}

Train the model on a 2D circle and try to
get a result of $x^\delta=y^\delta(\rho)=[cos(\rho),sin(\rho)]$,
and the corresponding differential equation is
$\frac{\partial^2 y_\delta}{{\partial \rho}^2}+y_\delta=0$
Result of the latent space structure shows in Figures\ref{fig:fig3} and Figure\ref{fig:fig4}, with the
numerical result $0.7360\frac{\partial^2 y_\delta}{{\partial \rho}^2}-0.0328\frac{\partial y_\delta}{{\partial \rho}}+  0.6761y_\delta=0$.


\section{Conclusion}
Summarize the key outcomes and potential future work.

Getting the data structure manifold in the
high-dimensional space is a hard task. In this article,
two methods were developed to get a dimension reduce
algorithm with greater ability to explanation.

The naive one
is to simply compute the PCA of neighborhood of each point
in the dataset, to find the latent differential structures
as shown in Figure\ref{one:dimensionalmani}, this can also
be treated as the global check if the data set meets the manifold hypothesis. With sufficient data, the first
method could basically learn all the nonlinear differential
relationships in the dataset, and take advantage of the ability of
PINN \cite[PINN]{raissi2017physics} to generate new data consistent
with the differential relationships.

The second method makes a linearity assumption, to significantly
reduce the requirement of the amount of data.
The augmentation of the Linearity assumption method could be
seen as a AutoEncoder who wants to find a parameterize agree
some Linear differential equations.With this method, the machine
refound the $sin$ and $cos$ function from the circle dataset.

For future work this model needs to be tested on more
complicated data set agree with more intricate differential relationships.
The Jacobian for high-dimensional latent space with high-order differential relationship
would require geometric amount of computational complexity $\sum_{j=0}^{N} D^{j}$ which may be augmented or
avoided by more advanced innovation in the future to have more efficiency.

\bibliographystyle{plain}  
\bibliography{references}  

\end{document}